\pdfoutput=1

\documentclass[11pt]{article}

\usepackage{ACL2023}

\usepackage{times}
\usepackage{latexsym}

\usepackage[T1]{fontenc}

\usepackage[utf8]{inputenc}

\usepackage{microtype}

\usepackage{inconsolata}

\usepackage{amsmath}
\usepackage{booktabs}
\usepackage{csquotes}
\usepackage{comment}
\usepackage{graphicx}
\usepackage{lipsum}
\usepackage{placeins}
\usepackage{pdfpages}
\usepackage{float}

\DeclareMathOperator*{\argmax}{argmax}

\graphicspath{ {./images/} }

\title{Language Model Sentence Completion with a Parser-Driven Rhetorical Control Method}

\author{Joshua Zingale\\
San Diego State University\\
5500 Campanile Drive\\
San Diego CA 92182 \\
\texttt{jzingale8274@sdsu.edu} \\\And
Jugal Kalita\\
University of Colorado Colorado Springs\\
1420 Austin Bluffs Pkwy\\
Colorado Springs CO 80918 \\
 \texttt{jkalita@uccs.edu} \\}

\begin{document}
\maketitle
\begin{abstract}
Controlled text generation (CTG) seeks to guide large language model (LLM) output to produce text that conforms to desired criteria. The current study presents a novel CTG algorithm that enforces adherence toward specific rhetorical relations in an LLM sentence-completion context by a parser-driven decoding scheme that requires no model fine-tuning. The method is validated both with automatic and human evaluation. The code is accessible on GitHub.\footnote{ https://github.com/joshua-zingale/plug-and-play-rst-ctg}\end{abstract}

\section{Introduction}

Despite outstanding success, Large Language Models (LLMs) are black-box in nature and perform unpredictably. They are known to generate non-facts and to deviate from desired criteria for generation \citep{ji_survey_2023}.
Controlled text generation (CTG) seeks to enforce constraints upon LLM-generated text, such as favoring the generation of pre-specified words or phrases or sentence structures, or requiring adherence to pre-specified communicative goals \citep{prabhumoye_exploring_2020}.

For a piece of text to be articulate, it must present a cohesive story using grammatically correct components that are also logically related to one another.
This paper presents a novel algorithm that attempts to influence the text generation behavior of an LLM by mandating that certain rhetorical relationships exist between spans of text.
The introduced algorithm incorporates a pre-existing parser that identifies discourse relationships among spans of text, within an LLM's probabilistic process of generation of text tokens, to produce text components that satisfy desired logical relationships.
In particular, given an input span of text, the approach generates the next span that holds a desired relation with the given input.

The direct use of the proposed system is the downstream task of generation of an entire Rhetorical Structure Theory (RST) tree. Such guided generation could aid specialized domains such as machine translation, where different languages have different expected orderings of rhetorical relations \citep{marcu_automatic_2000}.

The results show that the proposed method retains generation quality of an LLM, while enhancing it with a pronounced ability to control the rhetorical relations between adjacent sentence components.
Automatic and human evaluation verify the effectiveness of the control method in generating high-quality English text.

\begin{figure}[t]
\centering
\includegraphics[width=8.0cm]{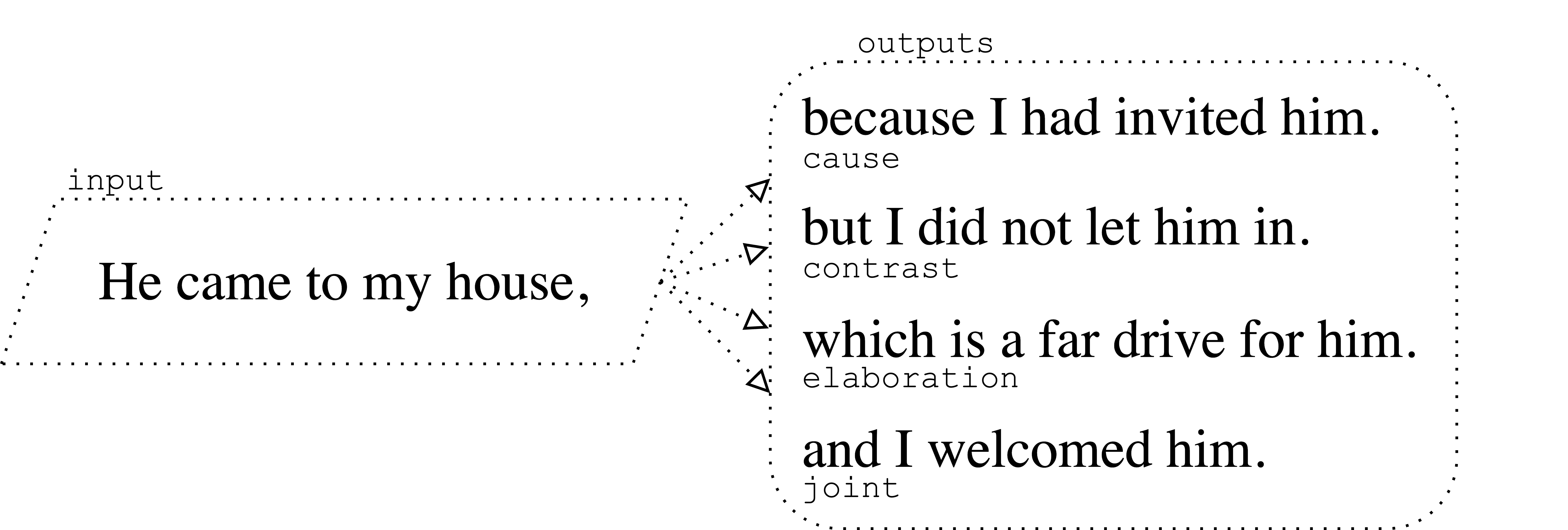}
\caption{Relation-influenced completions for the sentence, ``He came to my house,''. The proposed method generates such completions.}
\label{fig:exampleGenerations}
\end{figure}

\section{Related Work}
Large language models (LLMs), such as the GPT series \cite{openAI_2023},
PaLM \cite{narang2022pathways},
LLaMa \cite{meta2023introducing,meta2023introducing-2},
and BLOOM \cite{huggingface2022} generate text autoregressively, generating the next token conditioned on previously generated text. LLMs, pre-trained on vast corpora of text data, have demonstrated versatility in fluent text generation across domains \citep{wu_ai-generated_2023}.
However, these probabilistic models generate text in a black-box manner without the user's full understanding or control of the underlying generative process.
Controlled text generation attempts to modify the generation of text by LLMs by exerting influence on the next token being produced.

An attempt at controlling text generation includes
\citeauthor{baumler_hybrid_2022}'s use of phrase-structure parses of sentences generated by a language model and a database of world knowledge to modify generated phrases by inserting fact-driven words as applicable
(\citeyear{baumler_hybrid_2022}). \citet{zhou_think_2022} utilize a common-sense database to append knowledge to a language model prompt, enabling the language model to incorporate relevant information.
\citet{zhou_controlled_2023} use prompt engineering to instruct a language model to generate sentences with specific lexical, syntactic, semantic, style, or length constraints.

\citet{mann_rhetorical_1988} introduced a theory of discourse called Rhetorical Structure Theory formally to articulate how clausal units in a sentence and between sentences relate to one another to deliver meaning coherently. RST represents a collection of Elementary Discourse Units (EDUs) as a tree structure.
Although a descriptive theory, RST has been used to drive objectives in natural language processing, including summarization, machine translation, and generation \citep{afantenos_summarization_2005,marcu_automatic_2000,vander_linden_expressing_1995}. These early efforts in using RST to generate text were able to impose structure but were unable to generate fluent text.
On the flip side LLMs are good at generating fluent text, although are not amenable to being explicitly controlled.

The so-called plug-and-play approaches to CTG allow for controlled generation of text without expensive fine-tuning of the language models \citep{dathathri_plug_2020, zhang_survey_2023}. For example, \citet{liu_plug-and-play_2022} train a parser relevant to recipe generation and use it to re-rank the token distribution from a language model, resulting in controlled generation of recipes.

Building off the success of recent methods in integrating traditional computational linguistics tools, the present study integrates RST with large language modeling through a plug-and-play combination of an RST parser and a language model.

To the best of our knowledge, there is no equivalent method against which to test our system. We attempted to utilize prompting to guide BLOOM 1.7B toward generation of relation completions as a baseline;
but this smaller model showed no ability to complete these relations with prompt engineering.
This further bolsters the proposed method because, through it, the model can generate according to instructions that the model otherwise could not follow.

\begin{figure*}[tb]
    \centering
 \includegraphics[width=16cm]{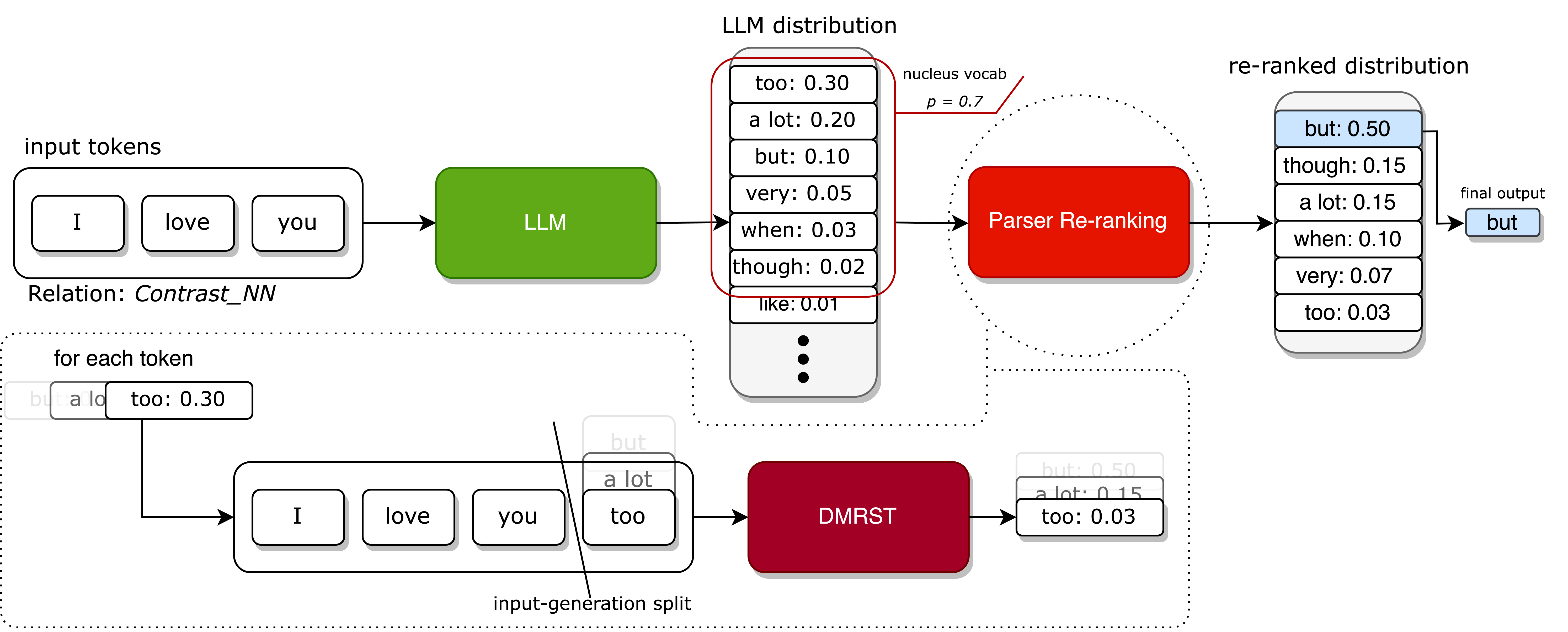}
\caption{The generation pipeline. Given the top-$p$ nucleus vocabulary of the distribution from the LLM, the parser re-ranks the tokens according to which tokens better fit the desired relation.
}
    \label{fig:generationPipeline}
\end{figure*}

\section{Models}
The proposed method uses two models for text generation. The first is a general language model without any RST pretraining. The second is an RST parser.

\paragraph{BLOOM 1.7B: }
BLOOM is a multilingual decoder-only transformer language model trained on the $1.61$ terabyte ROOTS corpus, which contains $46$ natural languages alongside $13$ programming languages \citep{workshop_bloom_2023,laurencon_bigscience_2022}.
The current study uses the 1.7-billion-parameter version of the model because of computational limitations for this study.
A BLOOM model is
decoder-only, allowing autoregressive generation of text.

\paragraph{DMRST:}
RST parsing consists of two tasks---segmentation and relation attribution.
Segmentation is the task of converting a document into a collection of EDUs,
the basic units in RST. Relation attribution, on the other hand, arranges these EDUs into a binary tree, assigning each edge to be a specific relation between two EDUs.
DMRST 
segments and parses raw text into an RST tree \citep{liu_dmrst_2021,liu_multilingual_2020}.
Importantly for the present study, DMRST also can be configured to perform relation attribution for a preset segmentation upon a document.

DMRST classifies between $42$ relations, where varying nuclearity configurations count as different relations. Each relation's name is of the form
\begin{align*}
\text{\{Relation\}\_\{Nuclearities\}},
\end{align*}
where \textit{Relation} is any of 18 categories, such as \textit{Contrast} or \textit{Attribution}, and \textit{Nuclearities} is \textit{NN} to mean the relation is between two nuclei, \textit{NS} to mean the left component is a nucleus and the right component is a satellite, and \textit{SN} for the other ordering of the nucleus and satellite.

The code for DMRST is publicly available.\footnote{https://github.com/seq-to-mind/DMRST\_Parser}

\section{Method}
Given a prompt and a relation, the pipeline generates a single EDU that continues the prompt while maintaining the given relation between the prompt and the generated EDU.
For each generation step, the language model first yields a distribution across all tokens conditioned on the prompt and the already generated tokens. Then, the RST parser re-ranks the top of the distribution to favor tokens that fit the desired relation.
Finally, the next token is selected from this re-ranked top of the distribution and the process continues until the parser detects the end of the EDU.

\paragraph{Generation:}
The pipeline receives relation $r$ and prompt $X$, comprising of a string of tokens, $x_1, x_2\ldots, x_U$, from the language model's vocabulary $V$. The pipeline then returns continuation $Y$, which comprises of tokens, $y_1, \ldots, y_T\in V$, such that $Y$ continues $X$ while maintaining relation $r$ with $X$.
Generation of token $y_t$ begins by finding the top-$p$, $0<p\leq1$, nucleus vocabulary $V^{(p)}\subset V$ \citep{holtzman_curious_2019}. $V^{(p)}$ is the smallest subset that satisfies
\begin{equation*}
\sum_{y\in V^{(p)}} P(y|X,Y_{<t}) \geq p,
\end{equation*}
where each token in $V^{({p})}$ is more likely than or equally likely to each token not in $V^{(p)}$, where $Y_{<t}$ is all tokens generated before timestep $t$, and where each $y$'s likelihood is calculated by the language model.
The size of $V^{(p)}$ is here capped at $k$.

The RST parser has token vocabulary $V'$, which is different from $V$. Therefore, the prompt and all tokens already generated are re-tokenized to $V'$ and are given by $X'$ and $Y_{<t}'$.
Each $y\in V^{(p)}$ is also re-tokenized to $V'$ and is given by $y'$, where $y'$ may be more than one token.

The RST parser then scores each $y\in V^{(p)}$ first by finding the logit value associated with the likelihood that the already generated sequence, $Y_{<t}'$, appended by $y'$, satisfies the desired relation $r$ with $X'$, calculated as
\begin{equation*}
\text{logit}_r(y) = D_r(X', Y_{<t}' \oplus y'),
\end{equation*}
where $\oplus$ is concatenation.
The DMRST parser
is given a preset segmentation such that the parser only finds the relation between $X'$ and $Y'_{<t} \oplus y'$.
After $\text{logit}_r(y)$ is found for each $y\in V^{(p)}$, the parser score for each $y$ is given by calculating a temperatured (with $\tau$) softmax function across all $\text{logit}_r(y)$:
\begin{equation*}
\text{score}_r(y) = \frac{e^{\frac{1}{\tau}\text{logit}_r(y)}}{\sum_{w\in V^{(p)}} e^{\frac{1}{\tau}\text{logit}_r(w)}}.
\end{equation*}

Following \citet{liu_plug-and-play_2022}, the next token, $y_t$, is calculated greedily with
\begin{equation*}
y_t = \argmax_{y\in V^{(p)}} [P(y|X,Y_{<t})^{(1 - \alpha)}\cdot \text{score}_r(y)^\alpha],
\end{equation*}
where $0\leq\alpha\leq1$ determines how much power the parser has to modify the language model's distribution and where, again, the likelihood of $y$ is provided by the language model.

\paragraph{Stopping: }
If the parser detects that an entire EDU has been generated, generation ends.

For ending generation, 
the DMRST segmenter is used.
Given an input string of tokens, the DMRST parser 
breaks up the string into EDUs.
For segmentation with the parser, we write, for some input sequence of tokens $W$,
\begin{equation*}
S(W) = (e_1, e_2,\ldots,e_L),
\end{equation*}
where $e_i$ is a sequence of tokens such that $e_i$ is itself an EDU and $e_1 \oplus e_2 \oplus\ldots\oplus e_L$ is the input sequence, $W$.
To know when to stop generation, the segmenter finds that the prompt, $X'$, has $P$ EDUs. Then, generation continues as outlined previously until the segmenter finds $S(X' \oplus Y_{<t}')$ to result in more than $P+1$ EDUs.
After stopping generation, the pipeline determines the smallest $N$ such that $X'\subset e_1\oplus e_2\oplus\ldots\oplus e_N$\footnote{$\subset$ here indicates a proper subset.}. The output, then, is $e_{1} \oplus e_{2}\oplus\ldots\oplus e_N$, with the input tokens, those from $X'$, removed from the beginning of the sequence.

\section{Experiments}

The proposed text generation method is evaluated both by automatic measures and by human feedback.
The method is tested with seven relations that were selected for their
ease of understanding to lay annotators.
Four volunteer native English speakers
each composed $20$ short English sentences according to instructions (Appendix \ref{sec:human-instructions}).
The instructions requested that the sentences be diverse in content, including tense.
The proposed method
generated eight completions for each of these $80$ sentences---seven for the seven relations being tested and one for no relation, that is, regular generation with the language model.

The parameter values used in the generation are $p = 0.75, k = 100, \tau = 0.1, \alpha = 0.7$.
For all completions, generation was forced, if it had not already stopped by itself, to cease after $30$ tokens or a period had been generated.

\begin{table}[tb]
\centering
\begin{tabular}{lccc}
\toprule Relation & \textbf{Correct\%} & \textbf{Perplexity} \\
\midrule
\midrule
\textbf{Cause\_NS} & 96.3 & 61.7\\
\textbf{Condition\_NS} & 58.8 & 44.1\\
\textbf{Contrast\_NN} & 95.0 & 52.4\\
\textbf{Elaboration\_NS} & 95.0 & 47.0\\
\textbf{Evaluation\_NS} & 33.8 & 56.2\\
\textbf{Joint\_NN} & 100 & 31.5\\
\textbf{Manner-Means\_NS} & 82.5 & 45.4\\
\textbf{All Relations} & 80.2 & 48.3\\
\textbf{None} & - & 43.9\\

\midrule
\end{tabular}
\caption{The automatic-evaluation statistics for each relation, where \textit{None} is generation with the language model alone.}
\label{table:automaticEvaluation}
\end{table}

\paragraph{Automatic Evaluation: }

The input text alongside its completion is automatically parsed using the DMRST parser.
As seen in Table \ref{table:automaticEvaluation}, five of the seven relations are parsed in accordance with each's desired relation more than $82\%$ of the time, four greater than or equal to $95\%$ of the time, and one is parsed to the desired relation for all tested prompts.
These results indicate that the proposed control method
effectively controls outputs such that they be parsed according to their desired relations.

Perplexity is used as a crude measure for the quality of the generated text, with lower numbers being better.
One worry
is that this control method may
degrade the quality of the generated completions. We therefore consider the average perplexity of completions generated without this control method for comparison.

Table \ref{table:automaticEvaluation} reveals that the secondary objective does not increase perplexity by much. In the case of \textit{Joint\_NN}, there even is a drop in perplexity from generation with no relation.
The results indicate that the control method does not cause the generated text to stray far from the language model's off-the-shelf distribution.
Thus, to the degree that BLOOM 1.7B accurately models language, the proposed method should also generate quality text.
Appendix \ref{sec:spanish-evaluation} has
similar automatic evaluation results for Spanish generations.

\begin{table}[tb]
\centering
\begin{tabular}{lccc}
\toprule Relation & \textbf{Rel} & \textbf{Flu} & \textbf{Rea}\\
\midrule
\midrule
\textbf{Cause\_NS} & 3.47 & 4.62 & 3.80\\
\textbf{Condition\_NS} & 3.25 & 3.82 & 3.98\\
\textbf{Contrast\_NN} & 3.97 & 4.02 & 3.67\\
\textbf{Elaboration\_NS} & 3.70 & 4.35 & 3.75\\
\textbf{Evaluation\_NS} & 2.47 & 3.97 & 3.75\\
\textbf{Joint\_NN} & 4.02 & 4.05 & 4.32\\
\textbf{Manner-Means\_NS} & 3.57 & 3.57 & 4.13\\
\textbf{All Relations} & 3.49 & 4.05 & 3.91\\
\textbf{None} & - & 4.16 & 3.80\\

\midrule
\end{tabular}
\caption{The human-evaluation statistics for each relation, where \textit{None} is generation with the language model alone. The metrics are \textit(Rel[ation-fit]), \textit(Flu[ency]), and \textit(Rea[sonableness]).
}
\label{table:humanEvaluation}
\end{table}

\paragraph{Human Evaluation: }
A subset containing 210 generated completions is used for human evaluation.
The random
subset
contains $20$ completions for each of the seven relations and $70$ completions with no enforced relation.

Three native English
speakers evaluated the generations across \textit{fluency}, \textit{reasonableness}, and \textit{relation-fit} according to instructions in Appendix \ref{sec:human-instructions}.
The annotators first rated the \textit{fluency} and \textit{reasonableness} and then rated the \textit{relation-fit} of each completion
because it does not reveal which relations influenced which completions, avoiding biasing annotator ratings.
For all metrics, each prompt-completion pair was rated on a scale form one to five.

\textit{Fluency} measures how grammatically correct a sentence is.
\textit{Reasonableness} measures how much sense a sentence makes.

Table \ref{table:humanEvaluation} shows the average ratings for each relation.
The average \textit{fluency} for all relations is only slightly lower than for no relation, $4.05$ against $4.16$, with the \textit{fluency} for different relations ranging from $3.57$ to $4.62$. The average \textit{reasonableness} for all relations is actually higher than that for no relation, $3.91$ against $3.80$.
\textit{relation-fit} is the degree to which the generation
satisfied the desired relation.

The average annotator rating of \textit{relation-fit} for generation with each of the relations is presented in Table \ref{table:humanEvaluation}.
The overall average, $3.49$, is well within the positive range. \textit{Evaluation\_NS} is unique in being poor, receiving an average of $2.47$.

\section{Perturbation Analysis}
\begin{figure}[tb]
    \centering
\includegraphics[width=8cm]{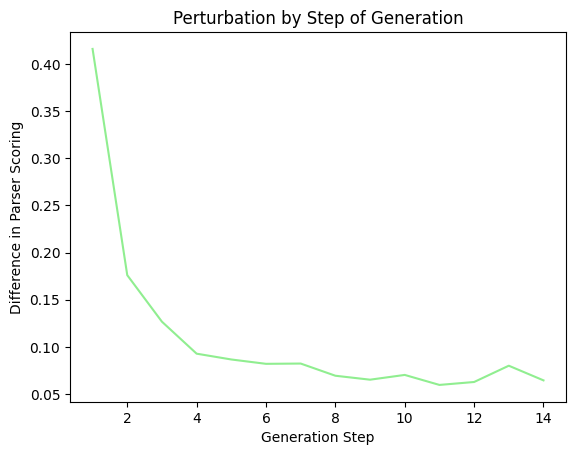}
\caption{At each step of generation, the average difference between the highest and the lowest DMRST parser-assigned score in the nucleus vocabulary across $560$ generations using seven different relations.
}
 \label{fig:perturbationGraph}
\end{figure}

Knowing where the proposed method most compels an alteration in token choice to occur grants insight to the problem of CTG with RST. We measure the degree of perturbation for each step of generation in a way semi-independent of $\alpha$, the generation parameter that determines how much the proposed method may perturb the language model's distribution.

After the top-$p$ nucleus vocabulary from the language model is obtained, the DMRST parser re-ranks each of these by creating a new token distribution, wherein each token is likely in as much as the parser sees the token to fit the desired relation.
The difference between the score of the highest and lowest parser-scored token is a proxy for how much the parser will re-rank, or perturb, the regular distribution.
When the difference is smaller, tokens are not re-ranked as much as when the difference is larger.
This, when only considering a single step of generation, is a measure independent of $\alpha$.

Figure \ref{fig:perturbationGraph} displays the average, across $560$ generations, of this difference for each generation step. The generations comprise of seven completions influenced by the relations heretofore used for each of the $80$ human-generated prompts.
Generation here used the same parameters as were used in Experiments. After the first token's generation, which has an average of $0.42$, the average difference drops to $0.18$ and
then after the fourth step below $0.1$. Hence, the most control is exerted during the generation of the first tokens,
which makes sense when considering that the words that explicitly begin the relation completions tested in this study for English are often headed with specific words or phrases. One example is \textit{Contrast\_NN}, for which completions typically begin with ``but'' or another adversative such as ``instead.''
After generating this first word or phrase, the decreased value of the difference, in conjunction with human evaluation confirming that
the proposed method maintains comparable fluency,
means that the language model, now generating conditioned on this initial relation-specific start, successfully adjusts to the desired relation without much further assistance from the parser.

\section{Conclusion}
Validated by automatic and human evaluation, the proposed control method is able to enforce a rhetorical relation during English sentence completion without sacrificing fluency or reasonableness.
The \textit{perplexity}, \textit{fluency}, and \textit{reasonableness} metrics show that the proposed method does not degrade the quality of generated text while \textit{correct\%} and \textit{relation-fit} indicates the control method's success.

\section{Limitations}
For lack of resources, the present study was not able to run reportable ablation studies with various generation parameters $p$, $k$, $\tau$, and $\alpha$. Also, the effect on the control method and on generation quality of replacing greedy generation, as was herein used, with sampling from the distribution or with beam search has not been measured.

For the human evaluation, there currently is no baseline for the \textit{relation-fit} metric, making the scores hard to interpret. Therefore the effect of the control method has been best measured here with the automatic scores, i.e. \textit{correct\%}. While this automatic metric does show that the control method conforms the language model to the parser, it does not guarantee that the generation's conforming to the parser indicates true completion of the controlled-generation task.

The proposed method requires that the a discourse parser be run between the prompt and generation for each of the considered next tokens. Thus is the computational overhead for generation increased.

\section{Acknowledgements}
All work herein reported is supported by the Nation Science Foundation under Grant No. 2050919.
Any opinion, finding, or conclusion in this study is that of the authors and does not necessarily reflect the views of the National Science Foundation.
We thank the participants that made the evaluation portion of this study possible.

\bibliographystyle{acl_natbib}
\bibliography{main}

\appendix

\appendix
\section*{Appendix}

\FloatBarrier

\renewcommand{\thesubsection}{\Alph{subsection}}

\subsection{Spanish Automatic Evaluation}
\label{sec:spanish-evaluation}

\begin{table}[tb]
\centering
\begin{tabular}{lccc}
\toprule Relation & \textbf{Correct\%} & \textbf{BLOOM} \\
\midrule
\midrule
\textbf{Cause\_NS} & 95.0 & 39.8\\
\textbf{Condition\_NS} & 43.0 & 25.2\\
\textbf{Contrast\_NN} & 99.0 & 31.3\\
\textbf{Elaboration\_NS} & 99.0 & 28.4\\
\textbf{Evaluation\_NS} & 36.0 & 26.1\\
\textbf{Joint\_NN} & 100 & 23.3\\
\textbf{Manner-Means\_NS} & 86.0 & 30.8\\
\textbf{All Relations} & 79.7 & 29.3\\
\textbf{None} & - & 19.5\\
\midrule
\end{tabular}
\caption{The Spanish-language automatic evaluation statistics for each relation, where \textit{None} is generation with the language model alone and \textit{All Relations} is all seven presented above combined. The same $100$ prompts are used to generate $100$ completions for each relation. \textit{Correct\%} is the percent of the generations that parse, using DMRST, to the relation that controlled their composition. \textit{BLOOM} is the generations' average perplexity as measured by BLOOM 1.7B.
}
\label{table:automaticEvaluationSpanish}
\end{table}

Since both BLOOM 1.7B and DMRST support Spanish, no modifications are needed for the system to work with Spanish.
Similar to the English automatic evaluation, we ran automatic evaluation on Spanish prompts.

To collect a set of Spanish-language prompts, ChatGPT 3.5 \citep{openAI_2023} was used to produce $100$ short diverse sentences in Spanish that employ various verb tenses.
As with the English prompts, the $100$ short sentences were converted to $100$ prompts by removing any trailing punctuation and adding a comma and a space where the punctuation was removed.

The same parameters as were used for the English generation are used to generate eight completions for each of the $100$ prompts---one for each of seven relations and one for no relation.
This leads to a total of $800$ Spanish completions.

Table \ref{table:automaticEvaluationSpanish} includes the same metrics as were used for English-language automatic evaluation.

As with the automatic evaluation for English, the proposed method effectively controls generation, i.e. is parsed to obtain the desired relation most of the time. $79.7\%$ of the completions result in the desired parsing. The method again does not increase the perplexity much, with an average relation perplexity of $29.3$ against the no relation perplexity of $19.5$. This again indicates that the method does not cause generation to stray far from the language model's regular distribution, implying that the quality of generation is comparable to that without the control method.

\subsection{Human Evaluation Instructions}
\label{sec:human-instructions}
Starting on the next page are attached the instructions given to the volunteers that generated the prompts for human evaluation and the instructions for the human annotators that rated the proposed method's generations.

\FloatBarrier
\clearpage



\section*{Supplementary material}

# COMPOSITION INSTRUCTIONS - sent to four volunteers

—

*volunteer_name*,

Please write 20 short sentences, each of which must integrate a specific motivation word. Make sure that the sentences are diverse in content and in verb use: 7 should be past tense (eg. was, had been), 8 present (eg. run), and 5 future (eg. will dive). Keep the structure of the sentences simple and try to write naturally. The motivation word may be used as a verb, noun, or otherwise in any sense of the word.

Write your sentences in this format:
Word: {motivation word}
{sentence integrating motivation word}
Word: {motivation word}
…

Here is an example submission:

Word: jump
The cat jumped onto the table.
Word: book
I am currently reading a fascinating book.
Word: park
Tomorrow, I will go for a jog in the park.
Word: set
The sun set over the horizon.
Word: funny
The baby giggled at the funny faces.
…


When you have written your 20 sentences, please email them to *researcher_email* in a format like the example submission above.

**Your motivation words** *(different list for each volunteer)***:**
Word: cluster

Word: board

Word: accept

Word: cupboard

Word: difficulty

Word: glacier

Word: cathedral

Word: cutting

Word: equal

Word: cat

Word: familiar

Word: presentation

Word: lunch

Word: cower

Word: wedding

Word: ritual

Word: limit

Word: industry

Word: sunshine

Word: candidate

# LABELING INSTRUCTIONS - sent to three paid annotators
—
You will be evaluating the quality of various English sentences by rating each sentence with a number 1-5 on three different metrics. First, evaluate the sentences in the *fluency and reasonableness* spreadsheet; then, evaluate those in the relationships spreadsheet. For both spreadsheets, do not linger on any question, but try and go quickly, letting your intuition guide your rating. The scoring can be subjective, so your own opinion should guide your scoring.

## FLUENCY & REASONABLENESS
In this spreadsheet, you will be rating the fluency and reasonableness of 210 sentences. For both scores, and for each sentence, you must provide a score of 1,2,3,4, or 5, with 1 meaning bad and 5 meaning good.

*Fluency* roughly measures how grammatically correct a sentence is. Grammatically correct here does not necessarily mean textbook grammar exclusively, but also informal grammar. For instance, "I ain't heard nothing" is fluent because a native English speaker may say it.

*Reasonableness* measures how much sense a sentence makes. A sentence like "I flew across the chair using a flip-flop" may be grammatically correct, but it is not reasonable. A reasonable sentence would be "I flew across the ocean using a plane."

Try not to conflate the *fluency* and *reasonableness* scores. For instance, if a sentence makes no logical sense but is fluent, the sentence should receive a high fluency; and if the sentence uses broken English but makes logical sense, it should receive a high reasonableness

## RELATIONSHIPS
In this spreadsheet, you will be rating the fittingness of 140 sentence completions to a specific relationship. You must provide a score of 1,2,3,4, or 5 for each sentence, with 1 meaning bad and 5 meaning good.

Each sentence is written as
*This is the first part of the sentence, (Relation) and this is the second part.* You must rate how well the second part of the sentence relates to the first part with the specified relation.

An example: *Tomorrow, I cannot wait to go to Disneyland, (Contrast_NN) but I am not looking forward to seeing my cousin Roger.* The second part of this sentence should contrast the first, which in this case it does, meaning that it

There are seven relations you will need to rate:
**Elaboration_NS**: The second part should elaborate, expand, or give more details concerning what the first part said. Eg. "I just missed my flight, (Elaboration_NS) which was to take me to Ohio"
**Contrast_NN**: The second part should contrast, contradict, or give an alternative to what the first part said. Eg. "I sent him a letter, (Contrast_NN) but I did not send one to his sister."
**Cause_NS**: The second part gives a cause, reason, or source of what the first part said. Eg. "I just missed my flight, (Cause_NS) because there was traffic on I-5."
**Manner-Means_NS**: The second part explains the manner in which, the means by which, or the way that an action in the first part is done. Eg. "I sent him a letter, (Manner-Means_NS) through the post-office."
**Evaluation_NS**: The second part evaluates, rates, or gives an opinion on the first part. Eg. "I just missed my flight, (Evaluation_NS) a major disappointment."
**Condition_NS**: The second part gives a precondition for the first part's action to hold. Eg. "I will go to the Moon, (Condition_NS) as long as you go too."
**Joint_NN**: The second part continues the first part without a strong rhetorical device. Eg. "I drive cars for a living, (Joint_NN) and have the loveliest wife in the world. "

You are not scoring the sentences for fluency or reasonableness, but for the quality of the relationship. If the second part relates to the first part as the relation says, then score it highly; if the second part does not relate to the first part as the relationship demands, score is lowly.

A bad example: "I went to the first floor, (Joint_NN) using the elevator." The second part does not relate to the first with a Joint_NN relation, so this should receive a low score.



\end{document}